\documentclass[10pt,twocolumn,a4paper]{article}

\usepackage[a4paper,margin=0.85in,columnsep=0.28in]{geometry}
\usepackage[T1]{fontenc}
\usepackage[utf8]{inputenc}
\usepackage{lmodern}
\usepackage{microtype}

\usepackage{graphicx}
\usepackage{amsmath,amssymb}
\usepackage{booktabs,array,multirow}
\usepackage{xcolor}
\usepackage[font=small,labelfont=bf,labelsep=period]{caption}

\usepackage[numbers,sort&compress]{natbib}
\usepackage[colorlinks=true,
            linkcolor=blue!60!black,
            citecolor=blue!60!black,
            urlcolor=magenta!70!black,
            pdftitle={VTLA-RL: Learning Visuotactile Policies for Load-Adaptive Grasping},
            pdfauthor={Qiwei Wu, Qixuan Li, Xin Xiang, Xiangyu Chen, Xiang Xiao, Daochen Shi, Ruihua Zhang, Renjing Xu}]{hyperref}

\newcommand{\todo}[1]{}

\title{\textbf{VTLA-RL: Learning Visuotactile Policies\\[2pt] for Load-Adaptive Grasping}}

\author{%
  Qiwei Wu\textsuperscript{1,$\dagger$}\quad
  Qixuan Li\textsuperscript{1,$\dagger$}\quad
  Xin Xiang\textsuperscript{1,$\dagger$}\quad
  Xiangyu Chen\textsuperscript{1}\\[5pt]
  Xiang Xiao\textsuperscript{2,$*$}\quad
  Daochen Shi\textsuperscript{2}\quad
  Ruihua Zhang\textsuperscript{2}\quad
  Renjing Xu\textsuperscript{1,$*$}\\[10pt]
  {\normalsize\textsuperscript{1}The Hong Kong University of Science and Technology (Guangzhou)\quad
  \textsuperscript{2}NVIDIA}\\[5pt]
  {\small\textsuperscript{$\dagger$}Equal contribution.\quad
  \textsuperscript{$*$}Corresponding authors.}
}

\date{}

\begin{document}
\maketitle

\begin{abstract}
Vision--language--action foundation models learn general manipulation priors, but object mass and local slip are difficult to infer from vision alone. Changing loads require contact-driven grasp adjustments that demonstrations collected under fixed conditions may not adequately cover. We present VTLA-RL to study visuotactile grasping policies and their reinforcement learning (RL) post-training under varying loads. Our main approach shares a visual encoder between tactile-image histories and RGB inputs to jointly generate and execute motion and bilateral force targets in a unified action space. Mixed real and simulated demonstrations first establish motion and contact priors. Two-stage multi-task RL then uses only simulated interaction to optimize the policy through task-success returns without continued real-demonstration supervision, and the trained policy deploys directly to real robots. We compare tactile representations and RL post-training routes in simulation and separately evaluate load variation and placement perturbations on hardware. The full method achieves higher grasping success than the supervised initialization, while RL post-training gains differ across approaches, providing empirical evidence for choosing representations and training approaches for this task.
\end{abstract}

\begin{figure*}[t]
\centering
\includegraphics[width=\textwidth]{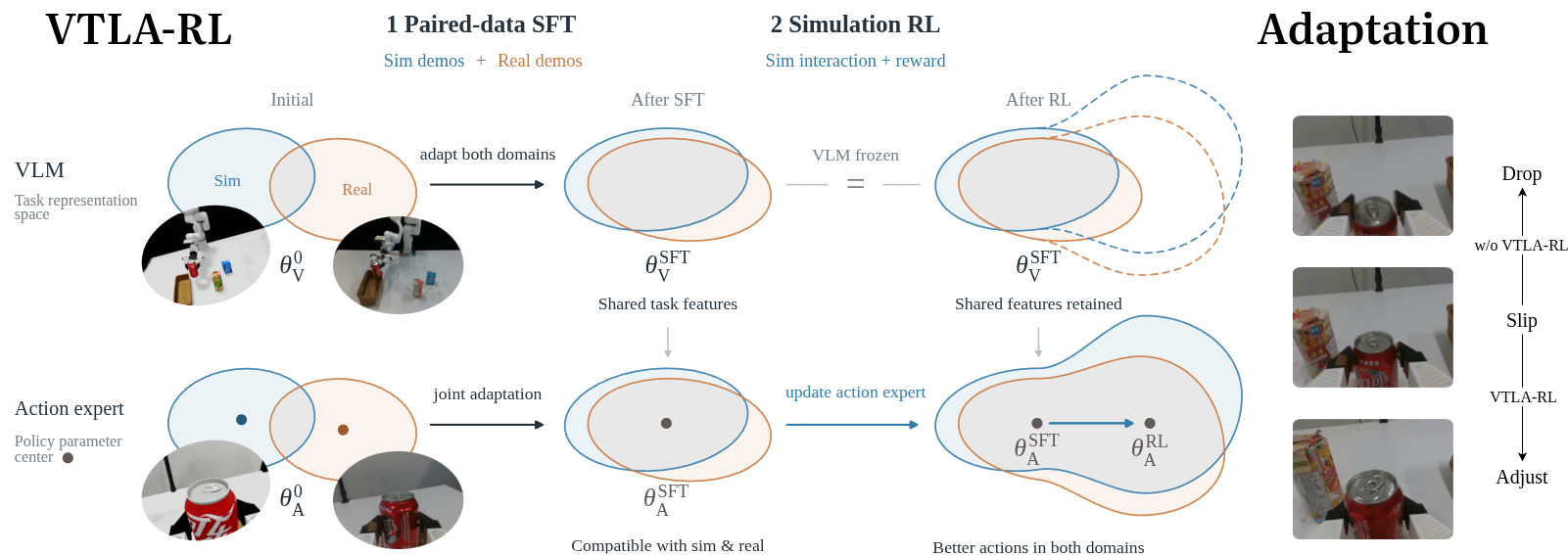}
\caption{\textbf{VTLA-RL overview.} After training on aligned real and simulated demonstrations, the vision--language--action (VLA) model produces relatively consistent actions across the two domains. During reinforcement learning (RL), we randomize object position and weight (friction) and train with task-success rewards. The trained policy responds appropriately to situations beyond the training set and performs well across a broader range of conditions. The right side shows one grasp-adjustment sequence following slip.}
\label{fig:overview}
\end{figure*}

\section{Introduction}
Moving a beverage carton from a table to a target area requires semantic understanding, geometric localization, and contact control. Vision can identify the target and plan motion, but has difficulty distinguishing visually identical containers with different internal masses before contact. Load changes alter the contact conditions required for stable grasping, and an initial grasp may slip during lifting or transport. To complete transport under varying loads, the policy must adjust its grasp pose, motion, and grip force based on the current contact state.

Vision--language--action (VLA) models provide transferable action priors; touch reveals contact deformation, slip tendencies, and force changes. Adding touch does not directly ensure that a policy can use this information to adjust grasps under changing loads. Imitation learning mainly inherits demonstrated motion and force habits, while demonstrations collected under fixed conditions may not cover new contact states. Simulated interaction offers a resettable training environment for expanding this experience and reduces the risks of drops, collisions, and material damage during hardware exploration.

We present VTLA-RL (Fig.~\ref{fig:overview}) to study visuotactile grasp adaptation under varying loads and the effects of tactile representations and reinforcement learning (RL) post-training routes. Without additional tactile pretraining, our main approach shares a pretrained visual encoder between tactile-image histories and RGB inputs, and uses one action expert to jointly generate motion and bilateral force targets. Supervised fine-tuning (SFT) on mixed real and simulated demonstrations establishes motion and contact priors. Two-stage multi-task RL in simulation then introduces placement and mass randomization in sequence to optimize the unified motion--force policy through task-success returns. The policy deploys directly to real robots through a hybrid force--position controller.

Starting from real and simulated demonstrations at fixed placements and masses, simulation RL expands interaction experience through placement and mass randomization. We evaluate grasping on hardware under placement perturbations and varying loads. Comparisons and ablations show higher success than the supervised initialization, with improved position generalization and load adaptation. Hardware results also show that the current tactile-array approach performs better at SFT initialization, whereas the image-based approach achieves larger post-training gains.

Our contributions are twofold:
\begin{itemize}
\item \textbf{A visuotactile policy for load-adaptive grasping.} We combine tactile history, a shared visual encoder, and a unified motion--force action interface with mixed-demonstration initialization and simulation RL to improve grasping without test-object mass labels. Post-training requires no continued real-demonstration supervision, and the policy deploys directly to real robots.
\item \textbf{Comparison of tactile representations and RL routes with real-robot evaluation.} We compare two tactile representations and three RL post-training routes in simulation under matched observation access, rewards, and interaction budgets, then separately evaluate load variation and placement perturbations on hardware. The experiments reveal differences in post-training gains and verify improved grasping performance of the full method over supervised initialization.
\end{itemize}

\section{Related work}

\subsection{Tactile representations and contact-aware foundation policies}
Sparsh learns tactile features across sensors through self-supervision~\cite{sparsh}; AnyTouch and AnyTouch~2 learn general representations of static/dynamic and force-related observations, respectively~\cite{anytouch,anytouch2}. TaF-VLA grounds tactile features in physical forces through tactile--force alignment~\cite{tafvla}. These approaches improve data coverage, sensor adaptation, and physical representation, but representation quality alone does not determine how much grip force a policy chooses in successful manipulation.

VTLA, Tactile-VLA, and VLA-Touch extend visual policies through preference learning, physical knowledge, and dual-level feedback~\cite{vtla,tactilevla,vlatouch}. OmniVTLA handles heterogeneous tactile inputs~\cite{omnivtla}; ForceVLA and ForceVLA2 emphasize force feedback, action generation, and closed-loop control~\cite{forcevla,forcevla2}; AT-VLA studies the timing and location of tactile injection~\cite{atvla}. DreamTacVLA and OmniVTA connect tactile prediction to contact actions~\cite{dreamtac,omnivta}. FTP-1 advances multi-source sensor pretraining~\cite{ftp1}; $N_0$-VTLA further uses large-scale tactile data and ALTER advantage-conditioned offline policy improvement~\cite{n0vtla}. ViTaL combines local visuotactile skills with residual reinforcement learning~\cite{vital}; TouchGuide uses touch to guide actions at inference~\cite{touchguide}; ViTaR selects and scales bounded residuals over a frozen VLA~\cite{vitar}; TouchWorld combines predicted tactile goals with high-frequency residual correction~\cite{touchworld}. HapticVLA uses safety-aware reward-weighted flow matching to penalize excessive grip force and distills tactile knowledge for deployment without tactile sensors~\cite{hapticvla}. VTLA-RL uses a Tabero-style hybrid force--position interface~\cite{tabero}, jointly generates motion and bilateral force targets, and optimizes them through simulation RL. With tactile feedback retained at deployment, it focuses on adapting motion and force to changing loads without object-mass labels.

\subsection{Tactile simulation and evaluation}
TACTO, Taxim, and FOTS generate optical tactile observations~\cite{tacto,taxim,fots}; TacSL incorporates tactile images and contact-force distributions into GPU learning~\cite{tacsl}; TacEx integrates soft-body contact and optical models through a modular framework~\cite{tacex}. DiffTactile supports differentiable contact and parameter calibration~\cite{difftactile}, Taccel targets high-fidelity contact with GPU parallelism~\cite{taccel}, TaCauchy provides continuum stress-based tactile simulation~\cite{tacauchy}, and TactileGenesis studies multiple tactile abstractions and their effects on policy learning~\cite{tactilegenesis}. Rich contact models reduce data cost, but introduce material parameters, numerical solving, and sensor-statistics matching.

UniVTAC unifies multisensor simulation, tactile representation, and manipulation benchmarks~\cite{univtac}; Tacmap reduces the domain gap through geometric contact representations~\cite{tacmap}. SoftVTBench separates task completion from object deformation and uses evaluator-only finite-element states to construct deformation-aware success, showing that contact quality requires independent evaluation~\cite{softvtbench}. VTLA-RL aligns optical tactile observations and motion--force control interfaces across simulation and hardware, then evaluates adaptation and transfer through grasping success under varying loads and placement perturbations.

\subsection{Reinforcement learning for VLA policies}
PiRL ($\pi_{\mathrm{RL}}$)~\cite{pirl} uses proximal policy optimization (PPO) to fine-tune the action expert of flow-based VLA policies. RL Token (RLT)~\cite{rlt} learns a compact readout representation, then freezes the VLA and representation while training a lightweight actor--critic. SimpleVLA-RL~\cite{simplevlarl} optimizes action-token VLA policies through RL, combining parallel trajectory sampling and exploration strategies to improve task performance. DSRL freezes the generative policy, treats the initial noise as a learnable action, and optimizes the noise input through RL~\cite{dsrl}. VTLA-RL's RL framework builds on PiRL and RLinf~\cite{rlinfpaper} and follows Hebero~\cite{hebero} for success-rate-weighted multi-task joint optimization.

TacCoRL~\cite{taccorl} injects normal-taxel histories into the VLA context and action expert through a dedicated encoder and contact gating, combining mixed-demonstration initialization with simulation PPO to improve recovery and completion in contact-rich tasks. Its main RL approach retains real-demonstration supervision. VTLA-RL focuses on grasping under changing loads, primarily uses tactile-image histories with the shared RGB encoder, and executes jointly generated motion and bilateral force targets through a Tabero-style hybrid force--position interface~\cite{tabero}. After mixed-demonstration initialization, RL post-training uses only simulated interaction and task-success returns, without continued real-demonstration supervision. We evaluate grasping adaptation under varying loads and placement perturbations.

\section{Method}
\subsection{VTLA}
\label{sec:tactile}
We use a $\pi_{0.5}$ backbone~\cite{pi05} with two $224\times224$ RGB views. TacImage concatenates eight tactile frames per finger, including the current frame, crops and downsamples to the RGB resolution, and shares the pretrained RGB vision transformer (ViT) without a separate tactile encoder. TacField shares TacCoRL's~\cite{taccorl} idea of encoding tactile-array histories as tokens; our causal temporal convolutional network (TCN) maps a reference frame and eight history frames of 2D marker arrays to one prefix token.

Each effective action has 13 dimensions: end-effector position (3), rotation vector (3), gripper (1), and contact-force targets (3 per finger: one normal and two tangential components, in newtons in its sensor frame). One flow action expert jointly generates motion and force targets, unnormalized for execution by a Tabero-style hybrid force--position controller~\cite{tabero}. Force targets receive demonstration supervision during initialization and remain action variables optimized by subsequent simulation RL.

\begin{figure*}[t]
\centering
\includegraphics[width=\textwidth]{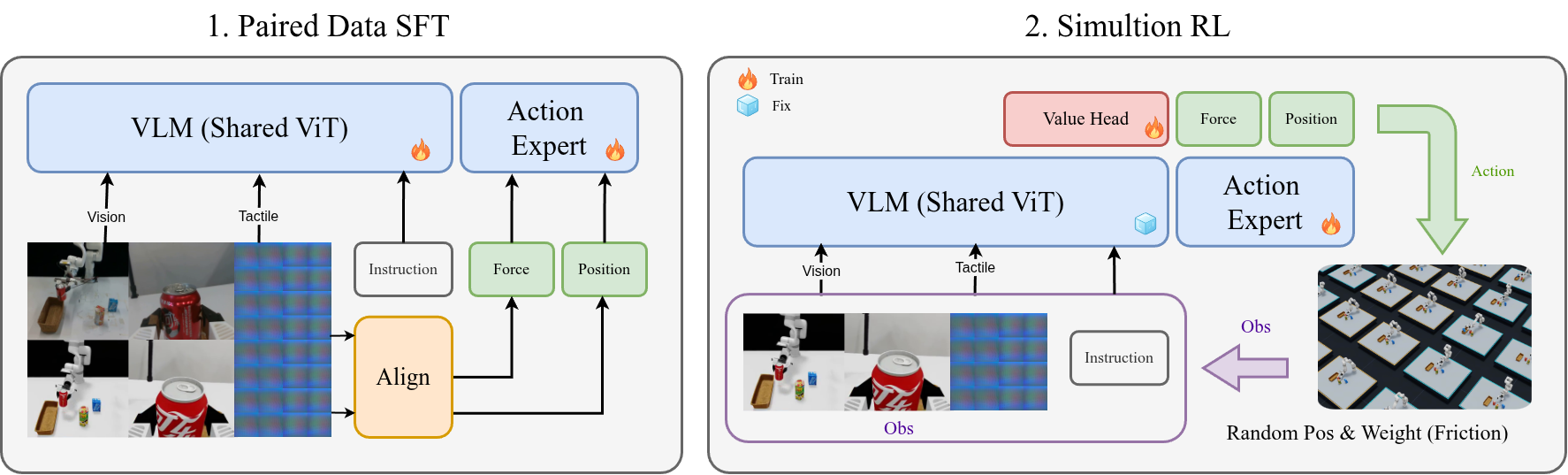}
\caption{\textbf{Policy architecture and RL loop.} Temporally concatenated tactile images and camera views share a vision transformer (ViT) encoder within the vision--language model (VLM). The flow-matching action expert jointly generates motion and force targets. Simulation rollouts provide data; RL updates action-expert low-rank adaptation (LoRA) parameters and the shared value head, while the visual encoder and remaining backbone remain frozen.}
\label{fig:model}
\end{figure*}

SFT initialization follows $\pi_{0.5}$'s flow-matching training~\cite{pi05}, jointly supervising motion and bilateral force targets in a unified action space to establish motion and contact priors. Normalized force coordinates receive separate loss weights to balance training scales; the appendix specifies the loss, weights, and trainable parameters. Figure~\ref{fig:model} shows the architecture.

\subsection{Sim-Real alignment}
\label{sec:data}
\begin{figure}[t]
\centering
\includegraphics[width=\linewidth]{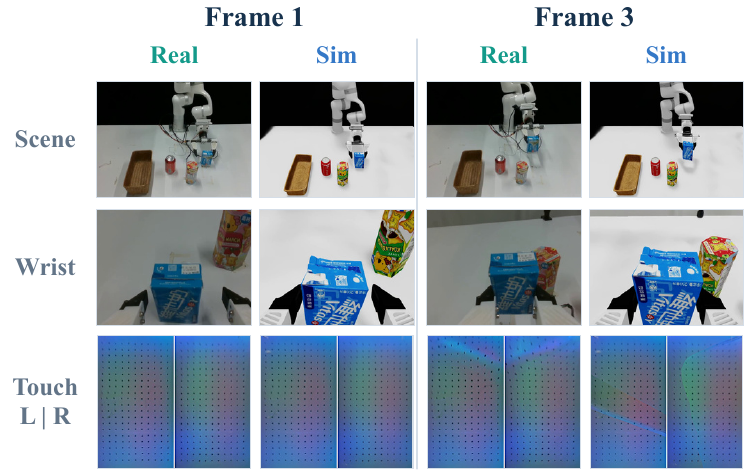}
\caption{\textbf{Sim--real correspondence.} Paired real and simulated external, wrist, and bilateral tactile views show milk-carton manipulation. Shared history windows, coordinates, and force units complete the policy interface alignment.}
\label{fig:alignment}
\end{figure}

Real and simulated scenes use corresponding geometry, cameras, and robots. Teleoperation supplies real demonstrations, and Tabero-style replay supplies simulated data. Three-dimensional Gaussian splatting (3DGS) reconstruction reduces visual differences. Figure~\ref{fig:alignment} shows visual and tactile alignment during grasping.

We combine Taxim~\cite{taxim} and FOTS~\cite{fots} in one sensor interface, with simulation assets built and adjusted for Xense sensors. Both use a shared contact representation from depth beneath the tactile pad. Taxim maps local surface geometry to tactile RGB images; FOTS combines contact deformation, centroid changes, and relative sensor--object yaw to produce image-coordinate marker displacements. These complementary observations of the same contact state consistently support tactile-image and array-history inputs.

Cross-domain calibration covers tactile observations and force control. We use real tactile calibration data to adjust the force-observation mapping. Across domains, we align image resolution, channel order, preprocessing, update timing, marker indexing, reference state, coordinates, history windows, and displacement scale. End-effector poses, gripper commands, and bilateral force targets share coordinate frames, units, and corresponding hybrid force--position execution interfaces, giving the actions in Sec.~\ref{sec:tactile} consistent physical meanings across domains.

We mix simulated and real demonstrations at a fixed 1:1 ratio using the SFT setup in Sec.~\ref{sec:tactile} to initialize motion--contact priors. Both subsequent RL stages use only simulation rollouts and task-success returns, without real-demonstration supervision (Sec.~\ref{sec:stagewise}). The policy deploys directly to real robots through the same observation/control interfaces, without additional fine-tuning, online real-robot RL, or test-object mass/friction labels.

\subsection{Multi-stage multi-task joint training}
\label{sec:stagewise}
Building on RLinf and PiRL~\cite{rlinfpaper,pirl}, we start from the mixed-data SFT checkpoint and use PPO~\cite{ppo} with generalized advantage estimation (GAE)~\cite{gae} to update action-expert low-rank adaptation (LoRA) parameters $\theta$ and value-head parameters $\phi$. The remaining backbone and tactile encoder stay frozen.

\paragraph{Multi-stage training.}
Stage 1 optimizes grasping and placement under randomized positions. Stage 2 resumes from this checkpoint and further randomizes object mass and friction, expanding interaction experience under varying loads and optimizing grasping behavior through task-success returns. Both stages share the policy, value head, and action interface, using sparse task-success reward: 1 on success and 0 otherwise.

\paragraph{Joint multi-task optimization.}
Tasks and their language instructions share one policy and value head. Following Hebero~\cite{hebero}, we adjust update contributions using task success. Let $s_i$ be the exponential moving average (EMA) of task $i$'s completed-episode success rate, with decay $0.9$, and $\bar s$ its equally weighted mean across tasks. The raw task weight is
\begin{equation}
\omega_i=\omega_{\min}+(\omega_{\max}-\omega_{\min})\sigma\!\left(\kappa(\bar s-s_i)\right).
\label{eq:taskweight}
\end{equation}
Here $\sigma(x)=1/(1+e^{-x})$, $\omega_{\min}=0.5$, $\omega_{\max}=2.0$, and $\kappa=10$; lower-success tasks receive larger weights. Raw weights are updated after each rollout and fixed during that round's optimization. Tasks with no completed episodes in the current round retain their previous success-rate EMA; all task weights remain one until every task is initialized. Let $\mathcal B$ contain a global optimizer batch's valid action-chunk samples. Each sample receives its task's raw weight, and these weights are normalized within the batch so that the sample weights $\hat\omega_k$ have mean one.

\paragraph{Model optimization workflow.}
Each round synchronizes training and inference parameters and collects up to 300 primitive steps per environment (30 fixed-length chunks). After aggregating success rates and updating task weights, we compute and fix GAE advantages and value targets from chunk rewards and old value estimates, then traverse valid samples in batches for one PPO epoch. Resets for the next round are prefetched during updating.

The policy predicts $H=50$ steps and executes only the first $K_{\mathrm{train}}=10$ during training. Of ten flow-denoising steps, one randomly selected step uses a stochastic differential equation (SDE) transition with noise level $0.5$; the others use deterministic integration. For chunk $k$, we evaluate the recorded stochastic transition's log density over valid executed steps and the 13 effective coordinates, grouped as
\begin{equation}
\ell_k(\theta)=\ell_k^{\mathrm{motion}}(\theta)+\ell_k^{\mathrm{force}}(\theta).
\end{equation}
The terms sum the first seven motion/gripper and final six bilateral-force coordinates from one action expert. With rollout parameters $\theta_{\mathrm{old}}$, evaluating both policies on the same recorded transition with the same valid-step mask gives $r_k(\theta)=\exp[\ell_k(\theta)-\ell_k(\theta_{\mathrm{old}})]$. This ratio concerns the selected stochastic transition. Averaging over the valid batch, the weighted PPO policy loss is
\begin{equation}
\begin{aligned}
&\mathcal L_{\mathrm{PPO}}=-\mathbb E_{k\in\mathcal B}\Bigl[\hat\omega_k\min\Bigl(r_k(\theta)\hat A_k,\\
&\quad\operatorname{clip}\!\bigl(r_k(\theta),1-\epsilon_{\mathrm{clip}},1+\epsilon_{\mathrm{clip}}\bigr)\hat A_k\Bigr)\Bigr].
\end{aligned}
\end{equation}
Here $\hat A_k$ is the precomputed chunk-level GAE advantage and $\epsilon_{\mathrm{clip}}=0.2$. The total loss is $\mathcal L_{\mathrm{RL}}=\mathcal L_{\mathrm{PPO}}+\mathcal L_V$, where $\mathcal L_V$ uses the same $\hat\omega_k$ and a clipped Huber loss to fit GAE value targets. Weights affect losses only, leaving rewards, advantages, and value targets unchanged. Motion and force share one ratio and task advantage, and both learn through task returns. SFT dimension weights do not enter RL; the main experiments add no real-demonstration or force-prediction supervision. Hyperparameters are in the appendix.

\subsection{Policy deployment}
\label{sec:exec}
Our custom hybrid force--position gripper reaches arbitrary openings at 20 Hz. A low-level controller following Tabero and Tactile-VLA tracks active targets using measured tactile force between policy updates. Synchronous and asynchronous execution both use the first 10 actions of each prediction at a 20-Hz base rate. Four interpolation points between targets yield 100-Hz commands. Model inference and arm execution run asynchronously, with tactile feedback correction at 20 Hz.

New chunks blend with existing fused targets using weights increasing linearly from $0.2$ to $0.8$ over three base steps; old chunks receive complementary weights. Rotations interpolate in local rotation-vector coordinates; other coordinates, including the gripper target, interpolate linearly. Recomputing fusion from original overlapping chunks after each admission avoids weight accumulation. We call this continuous-inference system with 20-Hz tactile feedback FAST and enable it by default.

\section{Experiments}
\label{sec:exp}
The experiments evaluate stable, load-adaptive grasping after VTLA-RL post-training and compare the effects of tactile representations and RL post-training routes. Real-robot experiments separately evaluate placement perturbations and varying loads.

\begin{figure}[t]
\centering
\includegraphics[width=\linewidth]{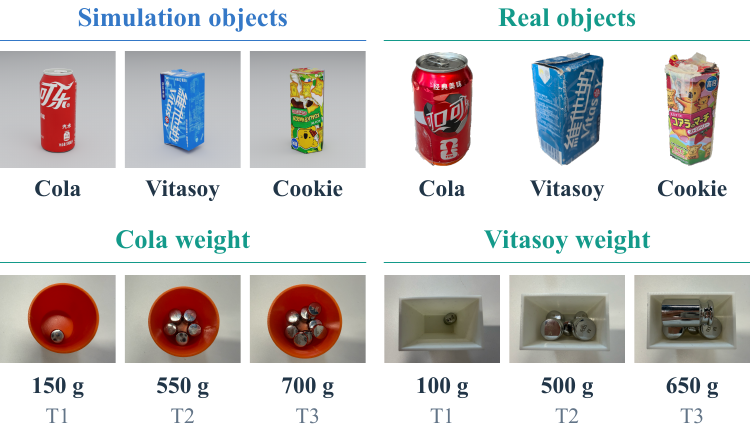}
\caption{\textbf{Test objects and mass conditions.} Top row: simulated Cola, Vitasoy, and Cookie, followed by their real counterparts. Bottom row: internal weight configurations for Cola at 150/550/700 g and Vitasoy at 100/500/650 g. Labels indicate total object mass.}
\label{fig:objects}
\end{figure}

\paragraph{Test tasks.}
The evaluation covers two tasks: a cylinder (Task A, a Coca-Cola can) and a cuboid (Task B, a milk carton). These are representative geometries. Data are collected with fixed object positions; XY perturbations are introduced during training, and the randomization range is further expanded at evaluation. In addition, we place a cookie box in the scene as a distractor object. During training, object mass is randomized over 150--250 g for Cola and 100--600 g for Vitasoy. At evaluation, we emulate unknown loads using three fixed task-specific tiers by changing the target object's mass: Task A (Coca-Cola) 150/200/250 g and Task B (Vitasoy) 100/300/600 g; the tiers are denoted T1/T2/T3, and friction is randomized within a bounded range. For the real-robot tests, to emphasize weight-induced slipping, we adjust the tiers to Task A 150/550/700 g and Task B 100/500/650 g; other settings follow those in simulation. Figure~\ref{fig:objects} shows the test objects.

\paragraph{Hardware and data sources.}
Training uses eight RTX PRO 6000D GPUs. Stage 0 performs SFT; the supervision data combine teleoperated real demonstrations with simulator replays of the same demonstrations, with 30 real trajectories and 30 simulated replay trajectories per task. RL post-training uses 128 parallel simulated environments, evenly split between the two tasks. Each of the two training stages runs for 200 epochs. Real evaluation uses an xArm with a tactile hybrid force--position gripper. Both real and simulated evaluation use the same RTX 4090.

\paragraph{Success criterion.}
Simulation success requires the object to remain inside the target container's goal region for eight consecutive control steps, subject to planar and height error requirements; the specific parameters are provided in the appendix. Episodes terminate after 300 control steps. Contact force does not participate in success. The real-robot criterion is essentially the same as in simulation.

\paragraph{Statistics.}
Each task is tested 10 times per random seed in simulation over five seeds, giving 50 trials per task; we report the mean and standard deviation of the success rate (SR). Success rates are reported in percent with the percent sign omitted. Each task is tested 10 times on real hardware, and we report the average success rate across all tasks. We also measure average grip force in newtons following Tabero~\cite{tabero}.

\paragraph{Hybrid force--position gripper.}
We use a hybrid force--position gripper and Xense tactile sensors. Unlike conventional open/close-command grippers, it responds at 20 Hz or higher at any opening degree and outputs grip force at the fingertips, with the hybrid controller realizing force adjustment. Details are given in the appendix.

\subsection{Weight-adaptive grasping under varying loads}
\label{sec:load}
We use Tabero's field-style model as the benchmark, initialized with the same $\pi_{0.5}$ pretrained parameters. Gentle and Firm undergo SFT following Tabero using demonstrations with different force commands, where the gentle force is about 50\% of the firm force. Adaptive undergoes SFT using data without force adverbs, followed by VTLA-RL's multi-stage training with the image representation of Sec.~\ref{sec:tactile}. Tables~\ref{tab:load-a} and~\ref{tab:load-b} report simulation results for the two tasks at three mass tiers within their respective training ranges. Gentle matches or exceeds Firm in some tiers, showing that stronger force commands do not always yield higher task success.

\begin{table}[t]
\caption{Task A results at 150/200/250 g; Gentle outperforms Firm at T3. Success rate (SR) is in percent without the percent sign; entries are mean (integer) $\pm$ standard deviation (one decimal) over five random seeds. Adaptive denotes our VTLA-RL method.}
\label{tab:load-a}
\centering
\begin{tabular}{lccc}
\toprule
Method & Tier 1 & Tier 2 & Tier 3\\
\midrule
Gentle (Tabero) & $98{\pm}4.5$ & $48{\pm}8.4$ & $32{\pm}8.4$\\
Firm (Tabero) & $98{\pm}4.5$ & $94{\pm}8.9$ & $18{\pm}16.4$\\
Adaptive (Ours) & $98{\pm}4.5$ & $98{\pm}4.5$ & $96{\pm}5.5$\\
\bottomrule
\end{tabular}
\end{table}

\begin{table}[t]
\caption{Task B results at 100/300/600 g; Gentle and Firm have equal success rates at T2. Protocol as in Table~\ref{tab:load-a}.}
\label{tab:load-b}
\centering
\begin{tabular}{lccc}
\toprule
Method & Tier 1 & Tier 2 & Tier 3\\
\midrule
Gentle (Tabero) & $74{\pm}24.1$ & $66{\pm}16.7$ & $64{\pm}5.5$\\
Firm (Tabero) & $90{\pm}7.1$ & $66{\pm}13.4$ & $76{\pm}11.4$\\
Adaptive (Ours) & $96{\pm}5.5$ & $94{\pm}5.5$ & $98{\pm}4.5$\\
\bottomrule
\end{tabular}
\end{table}

T1/T2/T3 denote task-specific mass tiers. Gentle and Firm perform differently across tiers: Gentle outperforms Firm on Cola at T3, while the two are comparable on Vitasoy at T2. Adaptive achieves high success across all tiers of both tasks.

\subsubsection{Average grip force analysis}
To examine whether RL post-training changes average grip force (AG), we measure AG in successful trajectories before and after RL for both tasks and both tactile representations, Img and Field. AG is the sum of the left- and right-sensor force measurements, averaged over valid samples throughout each trajectory. Table~\ref{tab:grip-before-after} shows small AG changes in the recorded successful trajectories, indicating that the observed success-rate gains are not accompanied by a large increase in this statistic. We hypothesize that the gains may reflect using force appropriately for specific geometries, rather than force increases alone.

\begin{table}[t]
\caption{Average grip force (AG, N) of different objects before and after RL. A and B denote Tasks A and B; Img and Field denote TacImage and TacField. Change is relative to Before RL.}
\label{tab:grip-before-after}
\centering
\begin{tabular}{lccc}
\toprule
Task & AG Before RL & AG After RL & Change\\
\midrule
A(Img) & 9.30 & 9.56 & $+2.8\%$\\
A(Field) & 8.74 & 8.98 & $+2.7\%$\\
B(Img) & 30.62 & 30.59 & $-0.1\%$\\
B(Field) & 31.14 & 30.54 & $-1.9\%$\\
\bottomrule
\end{tabular}
\end{table}

\begin{table*}[t]
\caption{Success rates of three RL routes and two tactile representations on two tasks and three load tiers.}
\label{tab:route}
\centering
\begin{tabular}{lcccccc}
\toprule
Method & A T1 & A T2 & A T3 & B T1 & B T2 & B T3\\
\midrule
VTLA-RL(Img) & $98{\pm}4.5$ & $98{\pm}4.5$ & $96{\pm}5.5$ & $96{\pm}5.5$ & $94{\pm}5.5$ & $98{\pm}4.5$\\
VTLA-RLT(Img) & $100{\pm}0.0$ & $100{\pm}0.0$ & $100{\pm}0.0$ & $78{\pm}16.4$ & $94{\pm}5.5$ & $92{\pm}8.4$\\
VTLA-DSRL(Img) & $98{\pm}4.5$ & $100{\pm}0.0$ & $98{\pm}4.5$ & $96{\pm}8.9$ & $82{\pm}16.4$ & $86{\pm}8.9$\\
VTLA-RL(Field) & $96{\pm}5.5$ & $94{\pm}5.5$ & $20{\pm}12.2$ & $86{\pm}5.5$ & $82{\pm}13.0$ & $80{\pm}10.0$\\
VTLA-RLT(Field) & $100{\pm}0.0$ & $92{\pm}4.5$ & $94{\pm}8.9$ & $4{\pm}5.5$ & $12{\pm}13.0$ & $6{\pm}5.5$\\
VTLA-DSRL(Field) & $94{\pm}8.9$ & $86{\pm}11.4$ & $0{\pm}0.0$ & $72{\pm}35.6$ & $60{\pm}7.1$ & $66{\pm}23.0$\\
\bottomrule
\end{tabular}
\end{table*}

\subsection{Method comparison}
We compare three RL post-training routes under the adaptive setting. VTLA-RL is the full method. VTLA-RL(Img) uses TacImage and matches Adaptive in Sec.~\ref{sec:load}. VTLA-RL(Field) uses tactile-array signals; its marker inputs and prefix-only TCN are specified in Sec.~\ref{sec:tactile}. VTLA-RLT corresponds to the RL Token route~\cite{rlt}: it freezes the VLA backbone and encoder, compresses prefix features into RL tokens, and trains a lightweight actor and value networks. VTLA-DSRL freezes the generator, treats the initial noise as a learnable action, and optimizes the noise input through RL~\cite{dsrl}. All methods use the same observation access, initialization data, two-stage RL configuration, reward, task weighting, execution rate, and primitive interaction budget; Table~\ref{tab:route} reports success rates in simulation for each method on the two tasks and three load tiers. Real-robot comparison results are reported in Sec.~\ref{sec:real}.

Equally averaging the six task--tier conditions in Table~\ref{tab:route}, VTLA-RL, VTLA-RLT, and VTLA-DSRL with Img achieve approximately 96.7\%, 94.0\%, and 93.3\% success, respectively. VTLA-RL(Img) has the highest overall mean and balanced performance across tasks. VTLA-RLT(Img) performs more consistently than VTLA-RL(Img) on Task A but trails it on Task B. Differences are larger with Field: VTLA-RLT(Field) performs well on Task A but poorly on Task B, while VTLA-RL(Field) and VTLA-DSRL(Field) also decline markedly at Task A's T3 tier. These results indicate that representation and post-training choices should be evaluated in relation to the task and load conditions in this setting. Section~\ref{sec:ablation} further compares RL post-training and execution, and Sec.~\ref{sec:real} reports hardware results.

\begin{figure*}[t]
\centering
\includegraphics[width=0.9\textwidth]{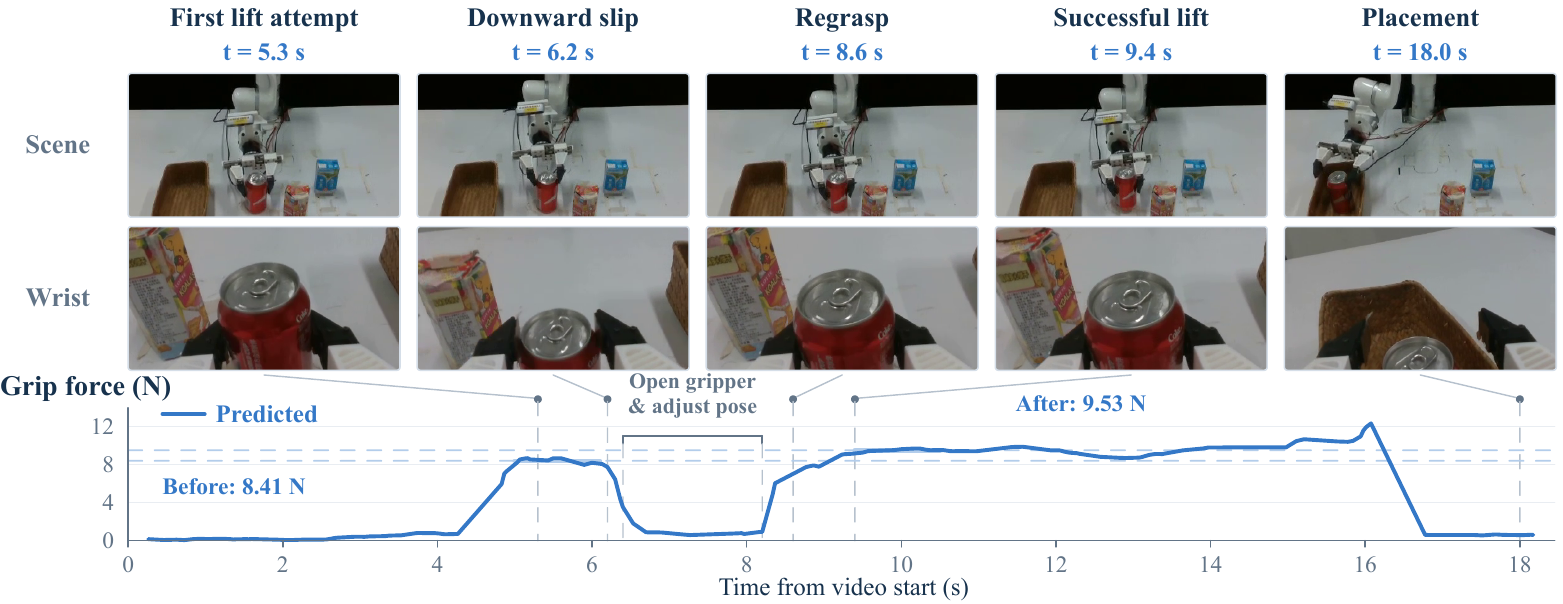}
\caption{\textbf{Real-robot grasp adjustment under slipping.} Synchronized scene and wrist views show initial grasp, downward slipping, grasp adjustment, and placement. The time axis preserves the recorded intervals between the selected keyframes. The blue curve represents the grip-force output predicted by the model.}
\label{fig:slipping}
\end{figure*}

\subsection{Ablation analysis}
\label{sec:ablation}
Table~\ref{tab:ablation} compares RL post-training and FAST for TacImage and TacField, reporting success rates at three load tiers and their average. The training-stage control compares the SFT checkpoint with the policy obtained by two-stage RL post-training from that checkpoint. The execution control compares asynchronous rolling inference with synchronous execution of 10-action prefixes on the same checkpoint. We also adapt PiRL to the same hybrid force--position action space, use the same joint, staged training setup, and remove tactile input to form the PiRL$^\dagger$ control. Table~\ref{tab:ablation} evaluates these controls in simulation; related real-robot comparisons appear in Sec.~\ref{sec:real}. For each tier, we first average the two tasks within each seed, then compute the mean and sample standard deviation across the five seeds. Avg. first aggregates the three tiers within each seed.

\begin{table}[t]
\caption{Module ablation: success rates (\%). Full uses FAST and RL; w/o removes the named component. $\dagger$Our force--position PiRL adaptation uses FAST and RL without tactile input.}
\label{tab:ablation}
\centering
\begin{tabular}{lcccc}
\toprule
Variant & T1 & T2 & T3 & Avg.\\
\midrule
\multicolumn{5}{l}{\textbf{Img}}\\
Full & $97{\pm}2.7$ & $96{\pm}2.2$ & $97{\pm}4.5$ & $97{\pm}2.4$\\
w/o FAST & $99{\pm}2.2$ & $98{\pm}2.7$ & $92{\pm}9.1$ & $96{\pm}2.5$\\
w/o RL & $90{\pm}7.9$ & $90{\pm}5.0$ & $70{\pm}6.1$ & $83{\pm}3.7$\\
\midrule
\multicolumn{5}{l}{\textbf{Field}}\\
Full & $91{\pm}5.5$ & $88{\pm}6.7$ & $50{\pm}7.9$ & $76{\pm}4.8$\\
w/o FAST & $86{\pm}12.4$ & $66{\pm}4.2$ & $36{\pm}9.6$ & $63{\pm}2.8$\\
w/o RL & $82{\pm}7.6$ & $63{\pm}9.1$ & $29{\pm}2.2$ & $58{\pm}4.3$\\
\midrule
PiRL$^\dagger$ & $80{\pm}9.4$ & $81{\pm}5.5$ & $62{\pm}13.5$ & $74{\pm}7.5$\\
\bottomrule
\end{tabular}
\end{table}

Table~\ref{tab:ablation} shows that RL post-training improves overall mean success from 83\% to 97\% for Img and from 58\% to 76\% for Field, benefiting both representations. Enabling FAST improves Img at T3 from 92\% to 97\%, with a small change in its overall mean across tiers, whereas Field's overall mean rises from 63\% to 76\%. Thus, execution has different effects across tactile approaches. Without tactile input, PiRL$^\dagger$ achieves an overall mean of 74\%, below Img Full and close to Field Full, showing that the effects of tactile approaches differ.

\subsection{Real-robot tests and method comparison}
\label{sec:real}
All models are deployed directly after simulation training, without additional real-world fine-tuning or online RL. Real demonstrations are used for SFT initialization, with no continued real-data supervision during simulation RL.

We first compare real-robot performance under fixed and randomized positions to assess transfer of simulation RL. Training details are given in the appendix. Img uses tactile images; Field uses array-style tactile force fields. Rows marked w/o RL use the mixed real--sim SFT initialization; the others undergo VTLA-RL training with a maximum position-randomization range of $\pm2$ cm. Fixed positions use the same placement as the data, and randomized positions apply a $\pm6$ cm planar perturbation to both objects in each test. Table~\ref{tab:real-pos} reports the average real-robot success rate over all tested tasks, with 20 trials per entry.

\begin{table}[t]
\caption{Real-robot success rates (SR, \%) under fixed and randomized positions.}
\label{tab:real-pos}
\centering
\begin{tabular}{lcc}
\toprule
Method & Fixed SR & Random SR\\
\midrule
Img & \textbf{100} & \textbf{65}\\
Img w/o RL & 65 & 35\\
Field & 95 & 55\\
Field w/o RL & 95 & 45\\
\bottomrule
\end{tabular}
\end{table}

Table~\ref{tab:real-pos} shows different hardware post-training gains for the two current representations. At SFT initialization, Field outperforms Img under both fixed and randomized positions. RL post-training improves Img from 65\% and 35\% to 100\% and 65\%, respectively, while Field changes from 95\% and 45\% to 95\% and 55\%. Thus, the representation ranking at supervised initialization does not persist after post-training under these conditions, and Img achieves larger success-rate gains. This may be because RL training helps the model better understand joint visual--tactile information represented as images.

We then test the models at the three object-mass tiers, keeping object positions fixed to control for placement. To reduce friction, we apply a film to the tactile sensors; otherwise, the high friction of the tactile sensors makes slipping rare, which hinders evaluating the policy's slipping and grasp-adjustment behavior. Table~\ref{tab:real-mass} reports the performance of the tactile-image VTLA-RL variants in the real-robot mass-tier tests, since the image-based variant already showed strong capability in Table~\ref{tab:real-pos}. VTLA-RL is the full method; VTLA-RLT and VTLA-DSRL correspond to the RL Token and DSRL post-training routes; VTLA-RL w/o FAST removes the asynchronous rolling inference speed optimization of Sec.~\ref{sec:exec} and uses synchronous execution of 10-action prefixes; VTLA-RL w/o RL is the mixed-data SFT initialization without simulation RL. PiRL$^\dagger$ follows the setting in Sec.~\ref{sec:ablation}. The table reports the average real-robot success rate over all tested objects, with 20 trials in total across all tasks per entry.

\begin{table}[t]
\caption{Real-robot method comparison across mass tiers. All tactile variants use TacImage (Img); the Img suffix is omitted. SR denotes success rate (\%).}
\label{tab:real-mass}
\centering
\begin{tabular}{lccc}
\toprule
Method & T1 SR & T2 SR & T3 SR\\
\midrule
VTLA-RL & \textbf{100} & \textbf{95} & \textbf{70}\\
VTLA-RLT & 45 & 45 & 30\\
VTLA-DSRL & 80 & 80 & 60\\
VTLA-RL w/o FAST & 85 & 75 & 60\\
VTLA-RL w/o RL & 65 & 65 & 35\\
PiRL$^\dagger$ & 90 & 50 & 25\\
\bottomrule
\end{tabular}
\end{table}

In Table~\ref{tab:real-mass}, the full method outperforms the RLT and DSRL routes at all three real-robot mass tiers and improves on SFT initialization by 35, 30, and 35 percentage points, respectively. At T3, the full method achieves 70\% success versus 25\% for PiRL$^\dagger$ without tactile input, supporting the role of touch in the current high-load setting; removing FAST yields 60\%. Figure~\ref{fig:slipping} qualitatively shows one grasping sequence for an object of unknown mass: after slipping, the policy opens the gripper, adjusts its pose, and regrasps before completing lifting and placement. The model's predicted grip-force targets also change during the sequence.

\section{Conclusion}
We presented VTLA-RL to study visuotactile grasping policies and their post-training under varying loads. The method combines mixed-demonstration initialization, a shared visual encoder, and a unified motion--force policy, followed by RL post-training through simulated interaction and direct real-robot deployment. Comparisons across tactile representations and RL methods in simulation and on real hardware, together with component ablations, demonstrate effective grasping under randomized placements and varying object masses.

\subsection*{Limitations}
The current method is evaluated only on small-scale data and tests, and the hardware experiments provide limited evidence for distinguishing small performance differences. Future work will expand task diversity and evaluation scale, reduce dependence on initialization, and improve feedback timeliness.

\clearpage
{
  \small
  \setlength{\bibsep}{1pt plus 1pt minus 1pt}
  \bibliographystyle{ieeenat_fullname}
  \bibliography{references}
}

\clearpage
\appendix
\setcounter{table}{0}
\setcounter{figure}{0}
\section*{Appendix}
\renewcommand{\thetable}{S\arabic{table}}
\renewcommand{\thefigure}{S\arabic{figure}}

\section{SFT Initialization and Loss}
\label{app:sft}
\paragraph{Data and initialization.}
We initialize from the PyTorch conversion of pretrained $\pi_{0.5}$. Each of the three SFT tasks, Vitasoy, Coca-Cola, and Cookie, contributes 30 real teleoperated demonstrations. Replaying the same actions in simulation yields 29, 30, and 30 usable trajectories, respectively, giving 179 trajectories and 26,741 frames in total. Real data recorded at 5 Hz are temporally aligned to 20 Hz. All usable trajectories are used for SFT. RL targets only Vitasoy and Coca-Cola and uses the simulated Cookie asset as a distractor. Paired replays provide observations of the same demonstrated actions in both domains.

Demonstrations use fixed placements and masses: 354 g for Coca-Cola and 57 g for Vitasoy. The main policy uses instructions without force adverbs, such as ``Pick up the Coca-Cola and put it into the basket.'' Gentle/Firm use instructions with the corresponding adverb.

\paragraph{Normalization and objective.}
Each label contains $H=50$ actions with 13 effective coordinates: six pose coordinates, one gripper command, and six bilateral force targets. Before normalization, the pose coordinates are expressed relative to the current state; gripper and force coordinates remain absolute. For each coordinate $d$,
\begin{equation}
\widetilde a_d=2\frac{x_d-q_{01,d}}{q_{99,d}-q_{01,d}+10^{-6}}-1.
\end{equation}
The training action sequences are flattened across time to estimate the 1st and 99th percentiles, using 5,000 histogram bins per coordinate. Normalized actions are zero-padded to 32 dimensions. Let $L_a$, $L_f$, and $L_p$ denote mean flow-matching squared errors over motion/gripper, force, and padding coordinates, respectively. The loss is
\begin{equation}
\mathcal L_{\mathrm{SFT}}=\frac{7L_a+6(0.05)L_f+19L_p}{32}.
\end{equation}
SFT updates rank-16 vision--language LoRA, rank-32 action-expert LoRA, the visual encoder, action input/output projections, and time MLP. TacField additionally trains its randomly initialized temporal encoder.

\section{Training Configuration and Workflow}
\label{app:training}
Table~\ref{tab:supp-stages} summarizes the training stages. SFT uses AdamW with $\beta=(0.9,0.95)$, weight decay $10^{-10}$, gradient-norm clipping at 1, and parameter EMA decay 0.99. The learning-rate schedule has peak/minimum values $2.5\times10^{-5}/2.5\times10^{-6}$, 10,000 warmup steps, and a cosine decay horizon of 1,000,000 steps.

\begin{table*}[t]
\caption{Stage configurations. RL rounds are outer rollout-and-update iterations.}
\label{tab:supp-stages}
\centering
\begin{tabular}{@{}>{\raggedright\arraybackslash}p{0.18\textwidth}>{\raggedright\arraybackslash}p{0.24\textwidth}>{\raggedright\arraybackslash}p{0.23\textwidth}>{\raggedright\arraybackslash}p{0.25\textwidth}@{}}
\toprule
Setting & SFT & RL stage 1 & RL stage 2\\
\midrule
Initialization & Pretrained $\pi_{0.5}$ & SFT step 30,000 & RL round 200\\
Length & 30,000 updates & 200 rounds & Continue to round 400\\
Objective/reward & Weighted flow matching & 1 on success, 0 otherwise & 1 on success, 0 otherwise\\
Placement & Fixed & Planar $\pm1$ cm & Planar $\pm1$ cm\\
Static friction & --- & 0.5 & Gripper: 0.5; others: 0.2--0.6\\
Global batch / seed & 32 / 42 & 384 / 42 & 384 / 42\\
\bottomrule
\end{tabular}
\end{table*}

The 128 parallel simulated environments are evenly split between the two RL tasks. Each round collects up to 300 control steps per environment, executing 10 steps per action chunk. Updates exclude invalid steps after episode termination. Training stops at fixed round counts, and RL policies are evaluated at round 400.

Task weighting uses success-rate EMA decay 0.9, weights in $[0.5,2.0]$, and sigmoid scale 10.

\section{Success Criteria and Evaluation Protocols}
\label{app:eval}
\begin{figure*}[t]
\centering
\includegraphics[width=\textwidth]{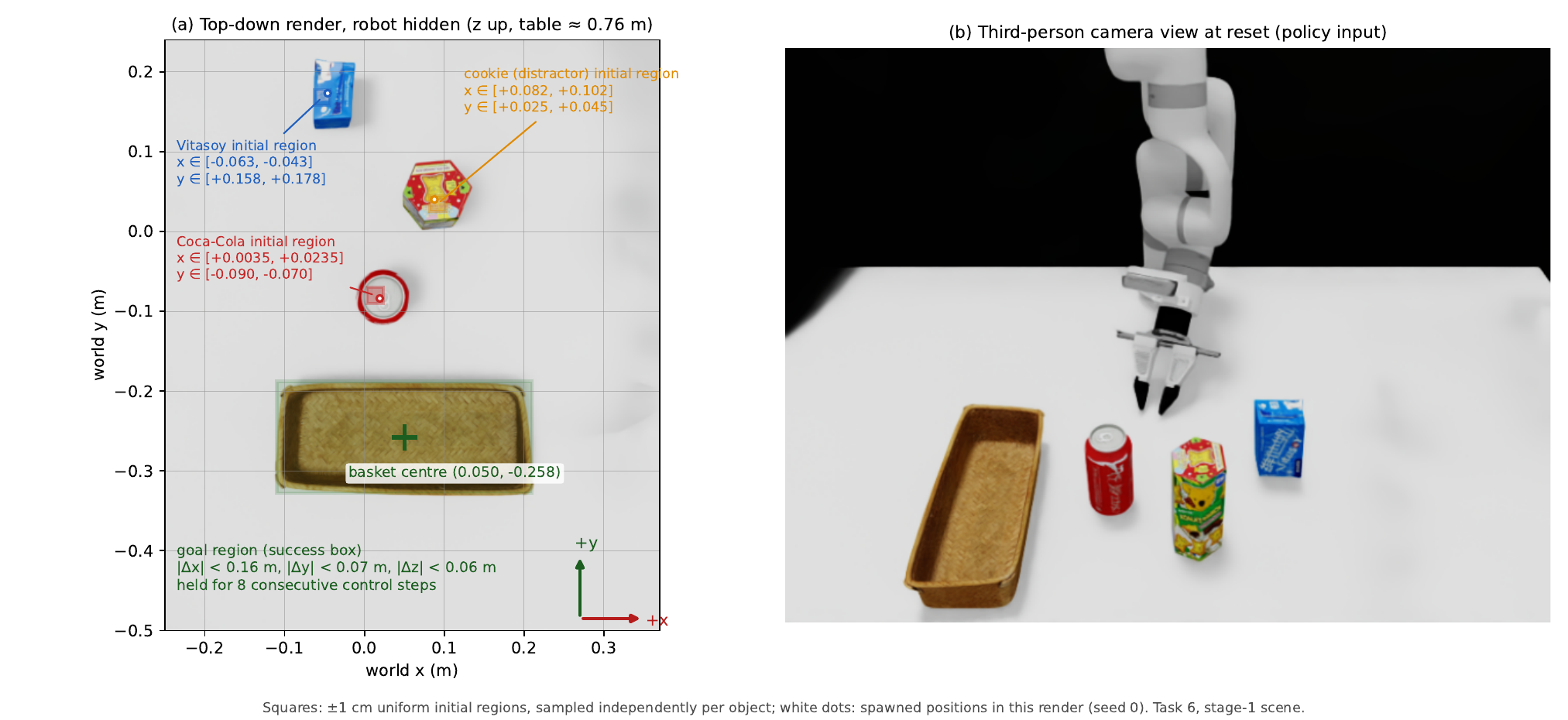}
\caption{\textbf{Goal region and camera view.} The green rectangle marks the horizontal success region. Colored squares indicate independently sampled $\pm1$ cm initial placement regions; the right panel shows the policy's third-person camera view.}
\label{fig:supp-goal}
\end{figure*}

\paragraph{Success and termination.}
Simulation uses a world frame with $z$ pointing upward. Relative to the basket center $(0.050,-0.258,0.761)$ m, the target must satisfy
\begin{equation}
|\Delta x|<0.16\,\mathrm m,\quad
|\Delta y|<0.07\,\mathrm m,\quad
|\Delta z|<0.06\,\mathrm m.
\end{equation}
Other objects must remain within 0.02 m and $15^\circ$ of their initial poses. These conditions must hold for eight consecutive control steps, or 0.4 s at 20 Hz. Force does not enter the success criterion. Figure~\ref{fig:supp-goal} shows the goal region.

Success terminates the episode. A triggered slip proxy or the 300-step (15 s) timeout counts as failure; simultaneous success and failure is counted as failure. Autonomous regrasping within an episode is allowed, with one outcome counted per episode. On hardware, placing the target in the basket with at most light contact with other objects counts as success. After a slip, the policy can still recover by adjusting and regrasping. In failed grasps of heavy objects, the gripper applied too little force to hold the object and the policy did not adjust its grasp. A timeout also counts as failure.

\paragraph{Sampling and statistics.}
Simulation uses seeds 160000--160004, with ten trials per task and seed, giving 50 trials per condition. The two graspable objects independently sample $x,y$ within $\pm1$ cm of their centers; orientations and the basket remain fixed. The gripper friction coefficient is 0.5, and other friction coefficients are randomized between 0.2 and 0.6. Success rates are summarized by their mean and sample standard deviation across seeds. For main-paper Table~5, tasks are averaged within each seed before computing statistics; its Avg. column also averages the three tiers within each seed. Hardware entries aggregate ten trials per task, or 20 in total.

\paragraph{Main-paper table settings.}
Tables~1--2 evaluate Coca-Cola at 150/200/250 g and Vitasoy at 100/300/600 g. Gentle/Firm use adverb-conditioned SFT; Adaptive uses TacImage PiRL at round 400. Table~3 compares successful trajectories from SFT step 30,000 and PiRL round 400 for both representations. Table~4 crosses the two tasks, three tiers, two representations, and three RL routes. Table~5 uses the same PiRL checkpoint for Full and w/o FAST, SFT for w/o RL, and a policy without tactile input for PiRL$^\dagger$.

Table~6 compares fixed placements with target-only planar offsets bounded by $\pm6$ cm, using SFT and PiRL for both representations. Table~7 uses fixed placements and film-covered tactile sensors: Coca-Cola at 150/550/700 g and Vitasoy at 100/500/650 g. Its tactile variants use TacImage. Evaluation is asynchronous except for w/o FAST.

\section{Gripper and Force Interface}
\label{app:control}
The platform combines an xArm 7, a custom single-actuator parallel gripper driven by a DM4310 motor, two Xense optical tactile sensors, and external/wrist RGB-D cameras. Gripper travel is 0--0.045 m, with a tool-center-point (TCP) offset of 0.232 m along link7's $z$ axis. Each pad measures $17.5\times29.3\times2$ mm and provides a $400\times700$ tactile image with an $11\times20$ marker grid.

Force targets and measurements use aligned local contact coordinates: $x$ points up, $y$ forward, and $z$ along the closing axis, with the two fingers' $z$ axes pointing in the same direction. Real forces are estimated by Xense from pad deformation. Simulated forces come from contact sensors and are transformed from world to fingertip coordinates; Taxim/FOTS supply tactile observations.

After unnormalization, the first six action coordinates add back the inference-time state to recover absolute TCP position and rotation vector. Simulation uses differential inverse kinematics; hardware converts TCP targets to flange poses before sending them through the robot SDK. The gripper coordinate sets opening, and the final six coordinates specify bilateral force targets.

The Tabero-style hybrid controller separates squeeze $s=|f_{L,z}|+|f_{R,z}|$ from net external load. A proportional force-error correction, with a deadband and travel limits, adjusts the predicted opening; end-effector translation regulates net load. Corrections are relative to each current prediction and do not accumulate across steps. The six force targets are realized through coupled arm/gripper motion, not six independent actuators; differences between the fingers' tangential forces have no independent force-driven rotation channel. Force feedback runs at 60 Hz in simulation and 20 Hz on hardware, with hardware commands sent at 100 Hz.

\section{Tactile Input Specifications}
\label{app:tactile}
\paragraph{TacImage.}
Eight frames per finger, including the current frame, are collected at 20 Hz. Each image is resized to $224\times224$ and then to a $56\times56$ tile, without cropping or background subtraction. A $4\times4$ grid places left-finger frames in the left half and right-finger frames in the right half, ordered from oldest to newest in row-major order within each half. The resulting $224\times224$ RGB image is normalized to $[-1,1]$ and shares the camera SigLIP encoder. With patch size 14, it contributes 256 visual tokens.

\paragraph{TacField.}
The input contains one fixed reference frame and eight recent frames of absolute marker coordinates, shaped $9\times440\times2$. The first 220 markers belong to the left finger and the remainder to the right, following fixed grid indices. Coordinates remain in pixels, without reference subtraction or additional normalization. Both representations repeat the earliest frame when fewer than eight history frames are available.

TacField uses two residual causal convolution blocks with kernel size 3, dilation 1, hidden dimension 4096, and SiLU activation, followed by a projection to 2048 dimensions. It has approximately 73.2M parameters and emits one token appended to the image/language prefix. Only the final time-step output is retained: its five-frame receptive field uses the five most recent frames, so the reference and earliest three history frames do not affect the output. The encoder is trained during SFT and frozen during RL. TacField shares TacCoRL's history-to-token idea, but uses marker coordinates and one prefix token, rather than taxel readings with tokens injected into both prefix and suffix.

\section{Sim--Real Alignment and Mixed Data}
\label{app:alignment}
Real and simulated data pair the same action trajectories with their respective observations. Real force preprocessing confirms jumps over four consecutive samples, with per-axis single-step thresholds $[4,10,4]$ N and an absolute cap of 60 N. Simulation uses reconstructed object assets and an xArm 7 asset with the same TCP offset, pose units, and rotation-vector convention.

The external camera is located at $(0.6993,0,1.5702)$ m in the world frame. The wrist camera is offset by $(0.11,0,0.05)$ m from link7 and rotated $-35^\circ$ about its local $x$ axis. Hardware uses the stored RealSense intrinsics; simulated cameras update at 20 Hz.

Taxim and FOTS share a tactile-pad height map from an internal depth camera. Taxim generates RGB images; FOTS combines the height map and relative sensor--object motion to produce 220 marker coordinates per finger. Both use sensor calibration with a 2 mm pad and a scale of 19.58 pixels/mm. FOTS normal-expansion, shear, and torsion coefficients are $(3.0,1.8,3.8)\times10^{-4}$. Grid parameters $(x_0,y_0,\Delta x,\Delta y)$ are $(29.8,50.0,32.8,31.75)$ for the left finger and $(31.95,56.0,33.7,31.45)$ for the right, in pixels. Gripper-pad static and dynamic friction coefficients are both 0.5.

\section{RL Routes and Training Budgets}
\label{app:routes}
Table~\ref{tab:supp-routes} summarizes PiRL, DSRL, and RL Token (RLT). The routes share environment count, outer rounds, control rate, and a ten-action training prefix. Nominal environment interaction budgets are matched. The routes differ in gradient updates, replay reuse, and the cost of learning the RLT representation. Proprioceptive state is available to critics only.

\begin{table*}[t]
\caption{RL implementation settings. Learning rates are listed as policy / value or critic. All routes freeze the VLA parameters outside their stated trainable modules.}
\label{tab:supp-routes}
\centering
\begin{tabular}{@{}>{\raggedright\arraybackslash}p{0.17\textwidth}>{\raggedright\arraybackslash}p{0.25\textwidth}>{\raggedright\arraybackslash}p{0.24\textwidth}>{\raggedright\arraybackslash}p{0.24\textwidth}@{}}
\toprule
Setting & VTLA-RL (PiRL) & VTLA-DSRL & VTLA-RLT\\
\midrule
Algorithm & PPO with GAE & SAC in initial-noise space & Off-policy actor--critic with BC regularization\\
Trainable modules & Action-expert LoRA (rank 32), value head & Noise actor, observation encoders, ten Q heads, entropy temperature & Residual MLP and twin Q networks, each $256\times3$\\
Action space & $10\times13$ executed coordinates & 32-dimensional initial noise, repeated across 50 steps & $10\times13$ normalized residual, bounded by $\pm4$\\
State access & No policy proprioception. The value head uses prefix features and proprioceptive state & No actor proprioception. The critic uses 7D proprioceptive state & The actor uses a 2048D RL token and reference actions. The critic uses the same token and 7D proprioceptive state\\
Exploration & One stochastic step among ten denoising steps; noise 0.5 & Gaussian noise policy; learned temperature, target entropy $-16$ & Fixed Gaussian standard deviation 0.05\\
Learning rates & $10^{-6}/10^{-4}$ & $10^{-4}/(3\times10^{-4})$ & $10^{-4}/(3\times10^{-4})$\\
Discount & 0.99; GAE $\lambda=0.95$ & 0.999 & 0.995\\
Update schedule & One pass per rollout & Eight updates per ready rollout & One critic update per 32 new transitions; critic:actor ratio 4:1\\
\bottomrule
\end{tabular}
\end{table*}

PiRL uses policy/value clipping of 0.2, Huber threshold 10, no advantage normalization, and no entropy bonus. DSRL averages its Q ensemble; RLT uses the minimum target Q. Both use target-network soft updates with coefficient 0.005. RLT additionally learns its 2048-dimensional token representation for 20,000 offline steps before RL, then freezes that encoder.

The DSRL actor uses two camera views plus the tactile image for TacImage, or two views plus an 880-dimensional tactile input for TacField. TacImage uses an 80,000-transition ring buffer; TacField uses trajectory replay with cache and sampling-window sizes of eight. RLT uses a 100,000-transition buffer and begins 2,000 warmup updates after collecting at least 256 transitions.

\begin{figure*}[t]
\centering
\includegraphics[width=\textwidth]{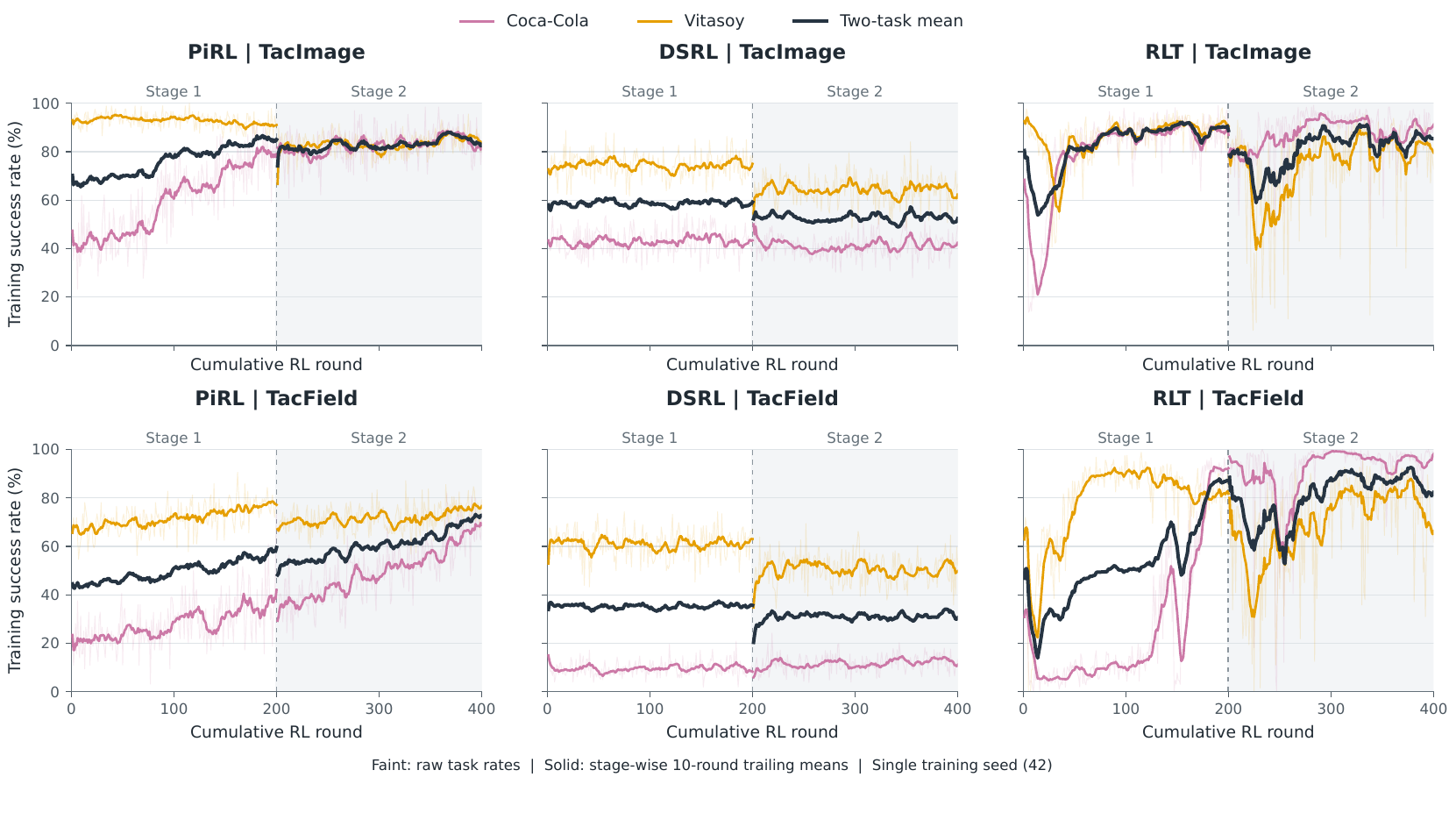}
\caption{\textbf{Training success rates in archived runs.} Each panel shows Coca-Cola, Vitasoy, and their equally weighted mean for seed 42. Faint lines show per-round task success rates, and solid lines show ten-round trailing means computed separately within each stage. Stage 2 starts after round 200. For RLT--TacField, rounds 211--212 use measurements recorded before a process restart.}
\label{fig:supp-training}
\end{figure*}

Figure~\ref{fig:supp-training} shows rollout success over rounds 1--400 in six archived runs conducted under a separate training configuration.
\begin{figure}[htbp]
\centering
\includegraphics[width=\columnwidth,trim=0 112bp 0 0,clip]{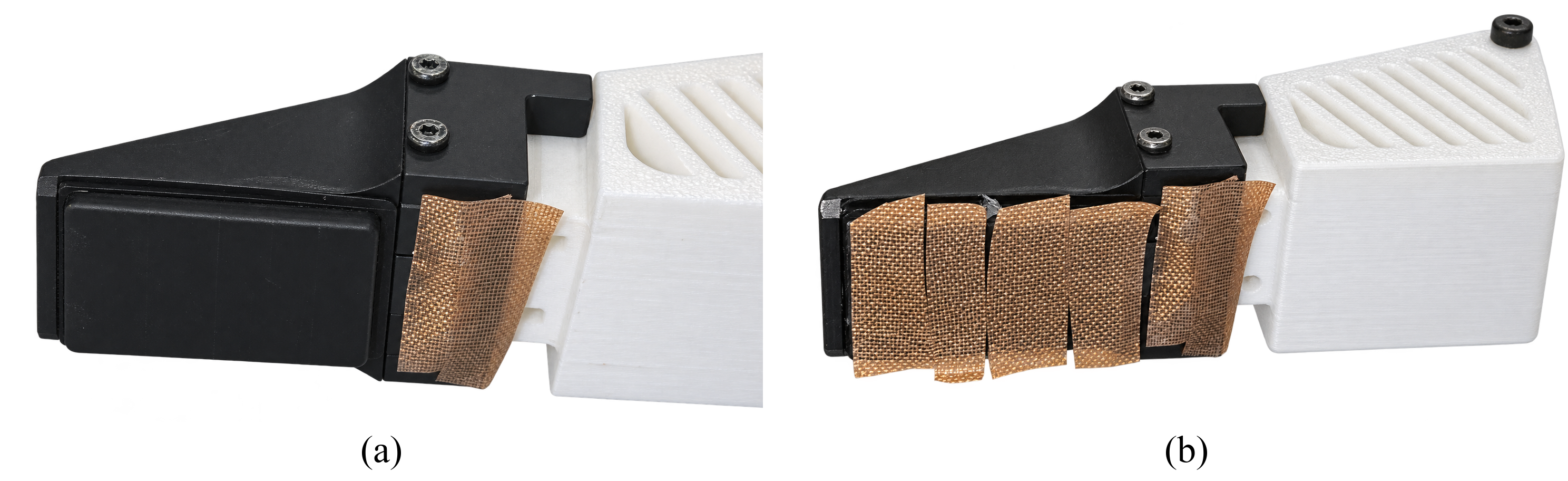}\par
{\footnotesize
\makebox[0.5\columnwidth]{(a)}\makebox[0.5\columnwidth]{(b)}\par
}
\caption{\textbf{Tactile sensor surface before and after applying the film.} (a) Bare tactile pad. (b) Tactile pad with the film used for slip evaluation.}
\label{fig:supp-sensor-friction}
\end{figure}

\section{FAST Timing and Ablation Definition}
\label{app:fast}
FAST runs observation collection, inference, and execution concurrently. Tactile history updates at 20 Hz. Predictions align to observation timestamps, and elapsed actions are skipped. Four interpolation points between base targets yield 100 Hz hardware commands. The force correction is applied after interpolation and removed from the action written back to the buffer, preventing accumulation.

Overlapping chunks are fused from oldest to newest. A new chunk's weight increases linearly from 0.2 to 0.8 over three base steps, with complementary weight on the previous result. Translation, gripper, and force interpolate linearly; rotation interpolates in local rotation-vector coordinates. Fusion is recomputed from original chunks after each admission.

Asynchronous simulation retains all 50 predicted actions (2.5 s) and requires admitted observation timestamps to be at least 0.2 s apart. Hardware retains the first 10 actions (0.5 s), admits returned predictions immediately, and permits one inference in flight. Its first chunk starts at arrival time. When waiting for a target, simulation pauses, with a 30 s timeout; hardware holds its last command, suspends force correction and history updates, and stops after 2 s. Fully expired hardware targets cause immediate stopping and are not replayed.

For TacImage PiRL at round 400, 40,594 simulation inference events give median/P95 wall-clock round-trip times of 0.353/0.953 s. The mean admitted-observation interval is 0.213 s in simulation time, approximately 4.7 Hz. These shared-service measurements distinguish wall-clock inference time from simulation-time refresh rate.

The w/o FAST control uses synchronous inference followed by ten open-loop actions. Simulation pauses during inference and refreshes the policy every 0.5 s; both execution modes retain the same 60 Hz force-feedback controller. Hardware w/o FAST also uses synchronous ten-action prefixes.

\section{Tactile Sensor Surface for Slip Evaluation}
\label{app:slip-surface}
Figure~\ref{fig:supp-sensor-friction} shows the tactile pad before and after applying the film used in the real-robot mass-tier tests (main paper, Table~7). The film covers the surface that contacts the object. Reducing friction at this interface makes slip events easier to observe during grasping and transport, allowing us to evaluate the policy's grasp adjustments.

\end{document}